\documentclass[11pt,a4paper]{article}
\usepackage[hyperref]{acl2021}
\usepackage{times}
\usepackage{latexsym}

\usepackage{microtype}

\aclfinalcopy 
\def\aclpaperid{46} 

\usepackage{tabularx} 
\usepackage{graphicx} 
\usepackage{url} 
\usepackage{amsthm,amsmath,amssymb} 
\usepackage{mathrsfs} 
\usepackage{bm} 
\usepackage{pifont} 
\usepackage{booktabs}
\usepackage{float}
\usepackage{lipsum}

\title{XRJL-HKUST at SemEval-2021 Task 4: WordNet-Enhanced Dual Multi-head Co-Attention for Reading Comprehension of Abstract Meaning}

\author{Yuxin Jiang*, Ziyi Shou*, Qijun Wang, Hao Wu, Fangzhen Lin \\
  HKUST-Xiaoi Robot Joint Lab on Machine Learning and Cognitive Reasoning \\
  Department of Computer Science and Engineering \\
  The Hong Kong University of Science and Technology, Clear Water Bay, Hong Kong \\
  \texttt{\{yjiangcm,zshou,qwangcd,hwubx\}@connect.ust.hk,flin@cse.ust.hk} \\}

\date{}

\begin{document}
\maketitle

\newcommand\blfootnote[1]{%
\begingroup
\renewcommand\thefootnote{}\footnote{#1}%
\addtocounter{footnote}{-1}%
\endgroup
}

\begin{abstract}
This paper presents our submitted system to SemEval 2021 Task 4: \textit{Reading Comprehension of Abstract Meaning}. 
Our system uses a large pre-trained language model as the encoder and an additional dual multi-head co-attention layer to strengthen the relationship between passages and question-answer pairs, following the current state-of-the-art model DUMA. The main difference is that we stack the passage-question and question-passage attention modules instead of calculating parallelly
to simulate re-considering process. We also add a layer normalization module to improve the performance of our model.
Furthermore, to incorporate our known knowledge about abstract concepts, we retrieve the definitions of candidate answers from WordNet and feed them to the model as extra inputs. 
Our system, called  \textbf{W}ord\textbf{N}et-enhanced \textbf{DU}al \textbf{M}ulti-head Co-\textbf{A}ttention (WN-DUMA), achieves 86.67\% and 89.99\% accuracy on the official blind test set of subtask 1 and subtask 2 respectively.
\end{abstract}

\section{Introduction}
Recently, there has been an increasing interest on Machine Reading Comprehension (MRC).\blfootnote{*Equal contribution.} 
While most MRC studies such as CNN/Daily Mail \cite{hermann2015teaching} focus on concrete concepts,
SemEval 2021 Task 4, \textbf{Re}ading \textbf{C}omprehension on \textbf{A}bstract \textbf{M}eaning (ReCAM),
targets abstract concept understanding, including \textit{imperceptibility} in subtask 1 and \textit{nonspecificity} in subtask 2.
The former, imperceptibility, highlights the abstract words that refer to ideas and concepts that do not
correspond directly to human perception. The latter is for hypernyms and abstract concepts such as
the class of vertebrate which includes whales as a concrete subclass \cite{changizi2008economically}.

\begin{figure}[tbp]
\centering 
\includegraphics[width=7.7cm]{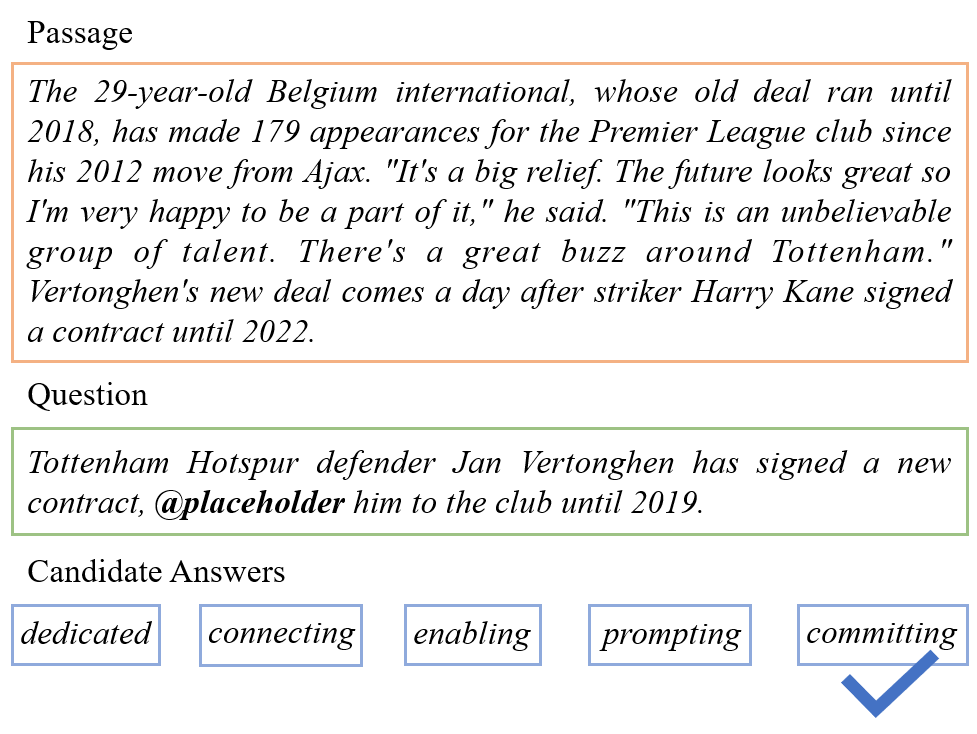}
\caption{An example of ReCAM subtask 1.} 
\label{question_example}
\end{figure}

In this task, given news fragments and incomplete abstracts, the machine needs to select the most suitable abstract words from candidate answers.
Figure \ref{question_example} shows one example of ReCAM subtask 1. 
Passage is the news sections and Question is a human written summary in which abstract words have been removed. 
Machines are requested to choose abstract words from five candidates for replacing \textit{@placeholder}.

For this shared task, we regard both subtasks as multi-choice MRC tasks. Various deep neural networks and attention mechanisms (e.g. \cite{dhingra2017gareader, wang2018comatch, zhang2019dual, zhang2020sgnet, jin2020mmm}) have been proposed to address these tasks.
In our work, following the state-of-the-art model DUMA \cite{zhu2020duma}, we adopt a Pre-trained Language Model (PrLM) as encoder and extend with an additional dual multi-head co-attention layer to strengthen the relationship between passages and question-answer pairs. 
For the dual multi-head attention layer, while DUMA builds passage-question and question-passage attention modules in a parallel way to simulate the transposition thinking process,
our model stacks two attention modules in order to simulate the process of re-considering for a deeper understanding of the passage.
More details on our attention calculation process can be found in Section \ref{sec:system}.
Furthermore, we add an additional layer normalization module immediately after the attention module.
From our experiments, we found that this additional normalization module definitely
improves our model's performance.

Our most significant design decision is to use WordNet \cite{miller1995wordnet} to expand on the abstract
concepts in the candidate answers. Intuitively, expanding an abstract concept according to its definition
in a dictionary should help as it helps relate the abstract concept with others that may occur in the text.
A key conclusion that we can draw from our experiments is that this is indeed the case.
One problem that we encountered when implementing this idea was that
most English words have multiple entries in WordNet. For example, \textit{bank} in WordNet can have
Noun definitions as well as Verb definitions.
We addressed this problem by using some heuristics and some additional information such as part-of-speech labels.
Because of the significant role played by WordNet, we call our system
\textbf{W}ord\textbf{N}et-enhanced \textbf{DU}al \textbf{M}ulti-head Co-\textbf{A}ttention (WN-DUMA).


We remark that 
our system did not use any additional training data for the tasks. In the final evaluation, our model is ranked 10 out of 23 and 9 out of 28 on the official subtask 1 and subtask 2  blind test set with 86.67\% and 89.99\% accuracy, respectively.
The code for our model is publicly available\footnote{\url{https://github.com/zzshou/RCAM}}.

The rest of the paper is organized as follows.  Section \ref{sec:system} gives the details of our system.
Section \ref{sec:experiment} describes our experimental setup including the datasets and hyper parameters used
for training.
Section \ref{sec:result} presents experimental results. Section \ref{sec:conclusion} concludes this paper with
some final remarks.

\section{System Description}

\begin{figure}[t]
\flushright 
\includegraphics[width=7.7cm]{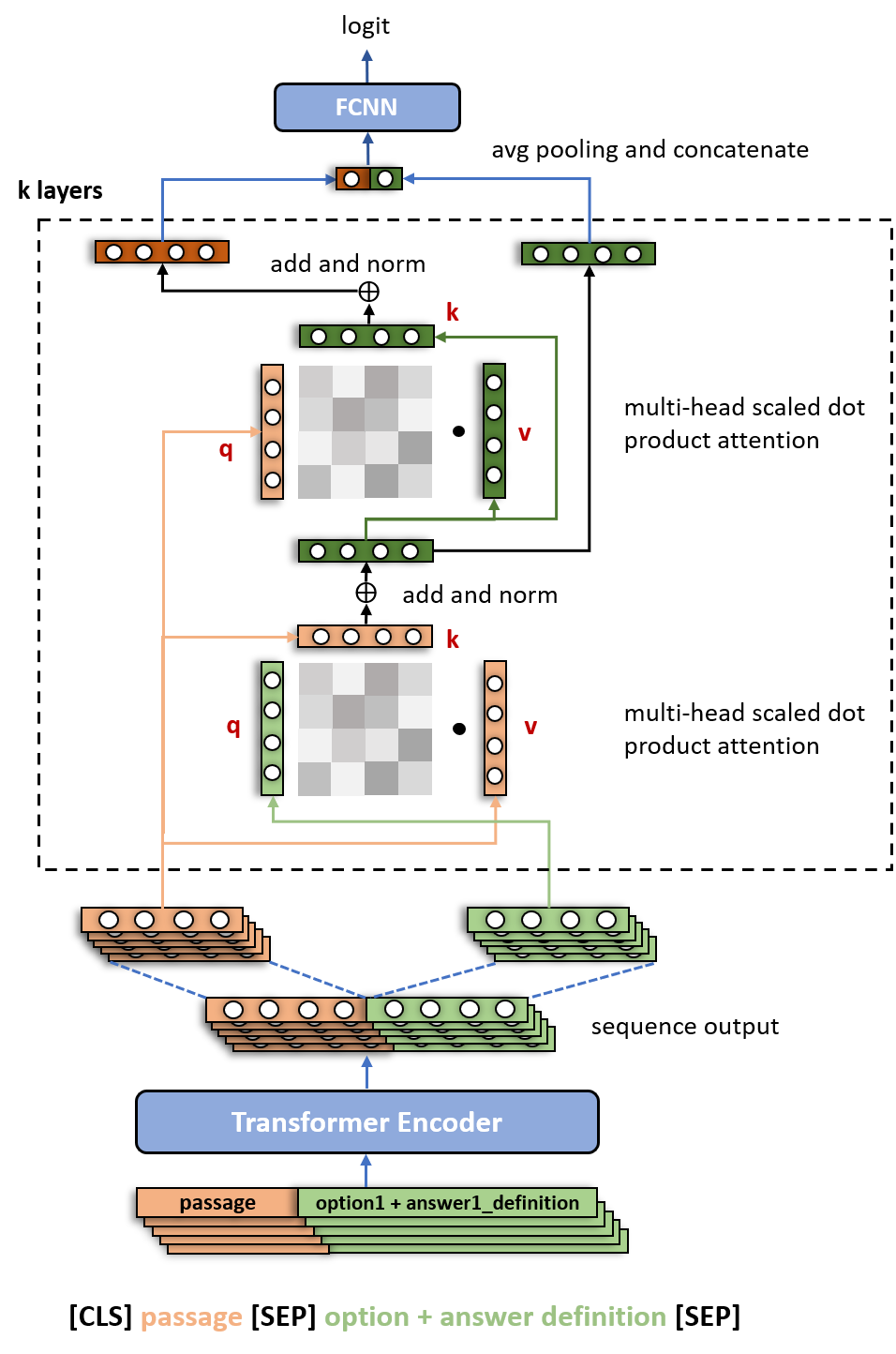}
\caption{The overall model architecture.} 
\label{model_architecture}
\end{figure}

In this section, We describe the framework of our end-to-end model
WN-DUMA.
Figure \ref{model_architecture} depicts the detailed architecture of our approach, with inputs at the bottom and outputs at the top.

\paragraph{WordNet-enhanced Encoder}
We regard both subtask 1 and 2 as multi-choice MRC problems. 
Such a problem includes a passage, a question with a \textit{@placeholder}, and 5 candidate answers to choose from.
First, we replace \textit{@placeholder} in the question with the given 5 candidate answers to form 5 options.
In the tasks, the candidate answers are all single words with abstract meanings, so we decided to 
add some extra knowledge from WordNet \cite{miller1995wordnet} to help the system better understanding the abstract meanings. 
More specifically, for a single candidate answer, we find its part-of-speech tag based on the option it's located in, and extract its definitions under this part-of-speech tag.
After tokenization, every instance is cast into the input form: [CLS] \emph{passage} [SEP] \emph{option + answer definition} [SEP]. 
To encode input tokens into representations, we feed them through a PrLM based on Transformer to obtain sequence embeddings, which draws a global relationship between the passage and the option-definition.

\paragraph{Dual Multi-head Co-Attention Layer}
Based on the above process, we further separate the output representations from transformer encoder to acquire the passage context embeddings $E^{P}\in \mathbb{R}^{d_{model}\times l_{p}}$ and the context embeddings of option-definition $E^{OD}\in \mathbb{R}^{d_{model}\times l_{od}}$, where $l_{p}, l_{od}$ denote the maximum length of passage and option-definition respectively.
Then based on the bi-directional matching network of DUMA which is quite similar to the multi-head self-attention module in vanilla transformer block \cite{vaswani2017transformer}, we first take $E^{OD}$ as Query, $E^{P}$ as Key and Value to calculate one of the co-attention representations, which simulates the process of human re-reading the passage with impression of option and definition. The formulas are listed as follows:
\begin{gather}
    Q_{i} = E^{OD}W_{i}^{Q}\\
    K_{i} = E^{P}W_{i}^{K}\\
    V_{i} = E^{P}W_{i}^{V}\\
    head_{i}=softmax(\frac{Q_{i}K_{i}^{T}}{\sqrt{d_{k}}})V_{i}\\
    MHA=Concat(head_{1},..., head_{h})W^{O}\\
    REP_{1}=Normalize(E^{OD} + MHA)
\end{gather}
where $W_{i}^{Q}\in \mathbb{R}^{d_{model}\times d_{q}}$, $W_{i}^{K}\in \mathbb{R}^{d_{model}\times d_{k}}$, $W_{i}^{V}\in \mathbb{R}^{d_{model}\times d_{v}}$, $W_{i}^{O}\in \mathbb{R}^{hd_{v}\times d_{model}}$ are linear transformations with learnable parameters, $d_{q},d_{k},d_{v}$ denote the dimension of \emph{Query}, \emph{Key} and \emph{Value}, $h$ denotes the number of heads.
Different from the structure of DUMA, here we make two changes: 1) apply "Add and Normalize" after getting the multi-head attention representation, which could result in more stable training. 
2) compute another co-attention representation by stacking rather than paralleling: take the acquired $REP_{1}$ as \emph{Key} and \emph{Value}, $E^{P}$ as \emph{Query}, which simulates the process of re-considering the option-definition with deeper understanding of the passage. 
Finally, we obtain $REP_{1}$ and $REP_{2}$, which have the same dimension as $E^{OD}$ and $E^{P}$, respectively. 
As a result, we can stack the co-attention module for k layers.

\paragraph{Classifier}
Here the co-attention representations $REP_{1}$ and $REP_{2}$ are merged and used for final classification:
\begin{gather}
    I_{1} = AvgPool(REP_{1})\\
    I_{2} = AvgPool(REP_{2})\\
    M = Concat(I_{1}, I_{2})\\
    logits = MW_{M}
\end{gather}
where $I_{1}, I_{2}\in \mathbb{R}^{d_{model}}$, $M\in \mathbb{R}^{2d_{model}}$, $W_{M}\in \mathbb{R}^{d_{model}\times n_{class}}$ denotes the one-layer fully-connected neural network, $n_{class}$ denotes the number of candidate answers. Consequently, for a single instance, we could get as many logits as the candidate answers, which are used to compute the corss-entropy loss by softmax.




\label{sec:system}

\section{Experimental Setup}
\label{sec:experiment}






\paragraph{Data and Metric} We used the official datasets \cite{zheng2021semevaltask4} provided by SemEval 2021 Task 4 competition. They were collected from BBC News in English language. 
Some basic statistics are listed in Table \ref{table1}.
\begin{table}
\centering  
\begin{tabular}{c|c|c}
\toprule
                       & Task 1 & Task 2 \\ \hline
Train                  & 3,227    & 3,318    \\
Dev                    & 837      & 851      \\
Test                   & 2,025    & 2,017    \\
Avg. passage length    & 270.3    & 429.7    \\
Avg. question length   & 24.6     & 27.1     \\
Vocabulary size        & 16,318   & 17,006   \\
Answer vocabulary size & 4,333    & 4,775   \\
\bottomrule
\end{tabular}
\caption{Basic statistics of subtask 1 and subtask 2 dataset.}  
\label{table1}  
\end{table}
According to the requirement of the organizers, participants could only use the corresponding dataset for a specific subtask to build models to ensure fairness. 
For better performance, technics like multi-task learning \cite{wan2020multitasklearning} are recommended for MRC tasks. 
In both subtask 1 and subtask 2, we utilize accuracy as the metric to evaluate our model performance. 

\paragraph{Hyper Parameters} All of our codes are written based on PyTorch\footnote{\url{https://pytorch.org/}}. 
To extract the word definition of candidate answers, we use NLTK toolkit \cite{BirdKleinLoper09}. 
The transformer encoder we used is pretrained ALBERT-xxlarge-v2 model\footnote{\url{https://github.com/huggingface/transformers}}. 
Since the code of DUMA is not open-source, we reimplement it by only using one co-attention layer where the attention heads are 64 and the dimension of \emph{Query}, \emph{Key} and \emph{Value} are all 64, because it is pointed that more co-attention layers do not improve the performance \cite{zhu2020duma}. 
The setting is also applied to our WN-DUMA for fair comparison.

Due to limited resources, the maximum sequence length of input tokens is set to 150 for both subtask 1 and subtask 2. 
In fact, we found that sequence length longer than 150 can only slightly improve the model performance. 
We choose mini-batch size equal to 2, and the AdamW optimizer \cite{loshchilov2017adamw} with an initial learning rate of 5e-06. 
We use some strategies for more stable training: 1) clip the gradient norm to 10; 2) adopt a linear scheduler with warm up of the first 10\% training steps. 
To avoid overfitting, we apply 0.1 dropout \cite{srivastava2014dropout} rate to the co-attention layer. 
We trained all the models for 3 epochs, evaluate on the dev set at every 200 training steps and save the model with the best dev accuracy. 
For each single model, we run experiments for 5 times with different random seeds and use the average as the ultimate performance.

\section{Results}
\label{sec:result}

\begin{table*}[h]
\centering  
\begin{tabular}{c|c|c|c|c}
\hline
Model                     & Task 1 dev   & Task 1 test & Task 2 dev   & Task 2 test \\ \hline \hline
$\rm BERT_{large}$ \cite{devlin2018bert}                     & 67.74          & -             & 69.45          & -             \\
$\rm RoBERTa _{large}$ \cite{liu2019roberta}                   & 74.31          & -             & 74.50          & -             \\
$\rm ALBERT_{xxlarge}$ \cite{lan2020albert}                   & 84.83          & -             & 82.84          & -             \\
$\rm ALBERT_{xxlarge}$ + DUMA \cite{zhu2020duma}              & 85.07          & -             & 86.13          & -             \\ \hline
$\rm ALBERT_{xxlarge}$ (question only)     & 79.57          & -             & 82.14          & -             \\ \hline
$\rm ALBERT_{xxlarge}$ + WN-DUMA (single)    & \textbf{85.90} & 84.54         & \textbf{87.43} & 86.61         \\
$\rm ALBERT_{xxlarge}$ + WN-DUMA (ensemble) & \textbf{87.57} & 86.67         & \textbf{90.01} & 89.99         \\ \hline
\end{tabular}
\caption{Model comparison on subtask 1 and subtask 2 dataset.}  
\label{table2}  
\end{table*}




\subsection{Quantitative Analysis}
Table \ref{table2} summarizes the experimental results. 
The first three models only have encoder (without enhancement of WordNet) and classifier part. It is clearly seen that ALBERT is much more efficient as encoder for abstract meaning understanding. 
It is worth noting that by only using the question and answer as input, the ALBERT model can also get pretty good results, as table \ref{table2} shows. 
Intuitively, it may be because the model could utilize syntax and semantics of the question sentence to choose the correct answer without looking through the passage.

Compared to $\rm ALBERT_{xxlarge}$, adding DUMA layer obtains around 0.2\% improvement in subtask 1, and more than 3\% improvement in subtask 2. 
Besides, our WN-DUMA single model achieves further improvements based on DUMA on both subtasks, +0.83\% and +1.3\% respectively, without increasing the number of parameters. 
Using a majority vote scheme, we ensemble our WN-DUMA model with different parameters for more stable predictions. 
Eventually, our ensemble models which get 87.57\% on subtask 1 dev set and 90.01\% on subtask 2 dev set acquire the best performance on test sets (86.67\% and 89.99\%, respectively) among our submissions.

\begin{figure}[t]
\centering
\includegraphics[width=8cm]{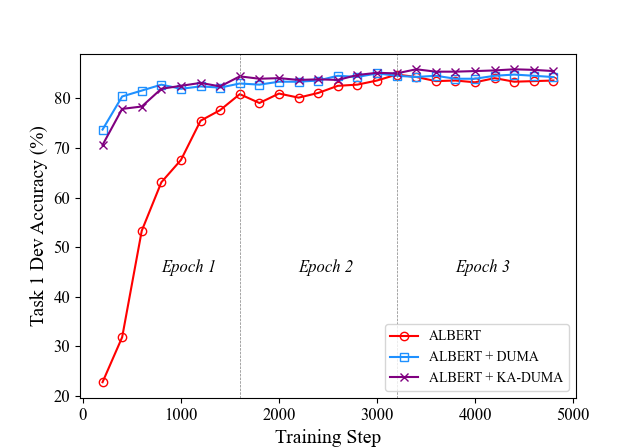}
\caption{Subtask 1 dev accuracy over number of training steps.} 
\label{task1_result}
\end{figure}

\begin{figure}[t]
\centering
\includegraphics[width=8cm]{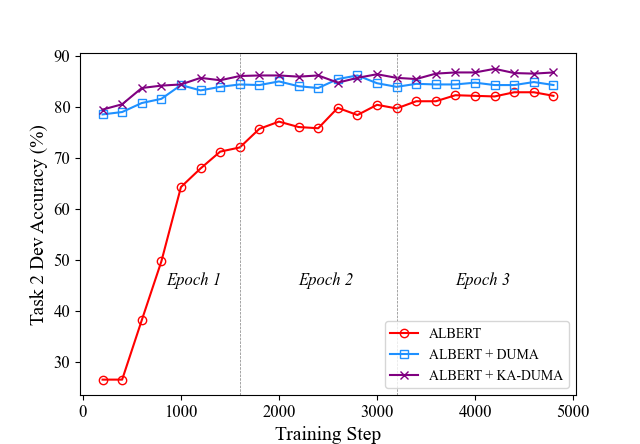}
\caption{Subtask 2 dev accuracy over number of training steps.} 
\label{task2_result}
\end{figure}

Figure \ref{task1_result} and Figure \ref{task2_result} illustrate the dev accuracy of different models on subtask 1 and subtask 2 as the number of training steps increases. 
It is interesting to observe that models with co-attention layer (DUMA and WN-DUMA) could get over 70\% accuracy with only 10\% of training examples.
While ALBERT model has to be trained with the full dataset to get relatively high accuracy.
Consequently, our WN-DUMA model may be useful when there only exists a small amount of training data.

\subsection{Error Analysis}
\begin{figure}[t]
\centering
\includegraphics[width=7.5cm]{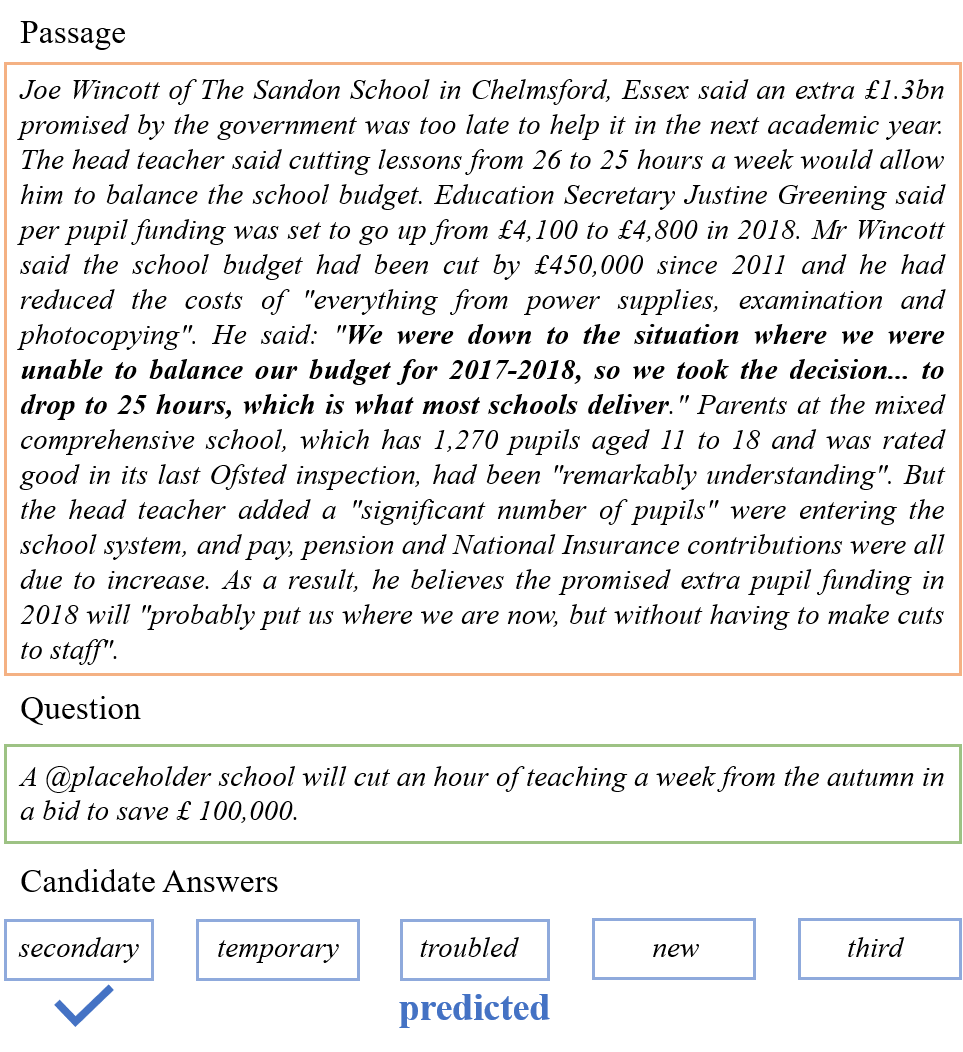}
\caption{An failure example made by WN-DUMA. The ground true is the answer with a correct mark at the bottom. While the prediction is the answer with "predicted" at its bottom.} 
\label{example2}
\end{figure}

In order to further improve our model performance in the future, we analyze some incorrect predictions made by WN-DUMA, and classify them into two categories:
\begin{itemize}
    \setlength\itemsep{-0px}
    \item Candidate answers with similar meanings. In some failure cases, the similarities between candidate answers are too high to distinguish. For example, outstanding and extraordinary, challenge and attempt, etc. 
    \item Lack of commonsense and relying too much on the information of the passage. Due to the fact that the question is the summary of the passage, the machine need to choose the most appropriate answer from a global perspective with some commonsense. However, our model make decisions by only capturing the local information in some cases. A specific example can be seen in Figure \ref{example2}. We can see that the model predicts that the answer is "troubled", most likely because the passage mentions "the school was trapped into financial difficulties". 
\end{itemize}

\section{Conclusion}
\label{sec:conclusion}
In this paper, we describe our submitted system in SemEval 2021 Task 4 ReCAM. Unlike previous MRC datasets, ReCAM focus more on machine's ability in understanding and representing abstract concepts. 
In order to provide more knowledge of abstract word, we extract WordNet definitions for each candidate answer based on part-of-speech tags.
In addition, our proposed WN-DUMA model consists of a PrLM as the encoder and a dual multi-head co-attention layer to enhance the relationship between passage and question-answer pairs as human's re-considering process.
Our WN-DUMA model improves the performance of our baseline model DUMA on these datasets.

There are some limitations in our experiments. 
Firstly, training data size of this task is limited compared to other MRC tasks, with less than 3400 training pairs in both subtasks.
This is understandable as collecting labeled data in many natural language processing tasks is expensive. 
Secondly,
using $\rm ALBERT_{xxlarge}$ PrLM, we only set 150 as the maximum text length in our experiments due to device limitation. 
Important sentences in the passage that are highly relevant to the summary are sometimes not covered. For PrLMs, their performance always improve as the number of their parameters increase. The use of large pre-trained models sometimes requires the sacrifice of context.
For our future work, we plan to explore ways to train models more efficiently with limited amount of labeled
data, and to design more cost-effective models to deal with long input texts.

\bibliographystyle{acl_natbib}
\bibliography{acl2021}


\end{document}